\documentclass[a4paper, 10pt, conference]{ieeeconf}

\newif\ifoagmfinalcopy

\oagmfinalcopytrue  

\IEEEoverridecommandlockouts                       

\usepackage{graphics} 
\usepackage{epsfig} 
\usepackage{mathptmx} 
\usepackage{times} 
\usepackage{amsmath} 
\usepackage{amssymb}  

\usepackage{lineno}
\usepackage{tikzpagenodes}
\usepackage{background}

\usepackage{multirow}
\usepackage{makecell}
\usepackage{cleveref}

\ifoagmfinalcopy
\backgroundsetup{color=white}
\else

\linenumbers

\newcommand{\MyOAGMConfidentialLogo}{
\begin{tikzpicture}[remember picture,overlay]
\node[align=center,text=blue] at ([yshift=1em]current page text area.north) {\Large \#\#\# ARW/OAGM 2021 SUBMISSION: CONFIDENTIAL REVIEW COPY \#\#\#};
\end{tikzpicture}%
}

\SetBgContents{\MyOAGMConfidentialLogo}
\SetBgPosition{current page.north west}
\SetBgOpacity{0.5}
\SetBgAngle{0.0}
\SetBgScale{1.0}

\fi

\title{\LARGE \bf
Real Estate Attribute Prediction \\ from Multiple Visual Modalities with Missing Data
}

\ifoagmfinalcopy
\author{Eric Stumpe$^{1}$, Miroslav Despotovic$^{2}$, Zedong Zhang$^{2}$ and Matthias Zeppelzauer$^{1}$
\thanks{$^{1}$E. Stumpe and M. Zeppelzauer are with the ICMT Institute of Creative Media Technologies, St. Pölten University of Applied Sciences, St. Pölten 3100, Lower Austria, Austria ({\tt\small estumpe@fhstp.ac.at; matthias.zeppelzauer@fhstp.ac.at})}  

\thanks{$^{2}$M. Despotovic and Z. Zhang are with the Kufstein University of Applied Sciences, Kufstein 6330, Tirol, Austria ({\tt\small miroslav.despotovic@fh-kufstein.ac.at; zedong.zhang@fh-kufstein.ac.at})}
\thanks{$^{3}$This research was funded by the Austrian Research Promotion Agency (FFG) project 880546 ``IMREA'' and we are grateful to DataScience Service GmbH for providing the data.}
}

\else
\author{Anon, Ymous}
\fi

\begin{document}

\maketitle


\begin{abstract}

The assessment and valuation of real estate requires large datasets with real estate information.
Unfortunately, real estate databases are usually sparse in practice, i.e., not for each property every important attribute is available.
In this paper, we study the potential of predicting high-level real estate attributes from visual data, specifically from two visual modalities, namely indoor (interior) and outdoor (facade) photos.
We design three models using different multimodal fusion strategies and evaluate them for three different use cases. 
Thereby, a particular challenge is to handle missing modalities. 
We evaluate different fusion strategies, present baselines for the different prediction tasks, and find that enriching the training data with additional incomplete samples can lead to an improvement in prediction accuracy.
Furthermore, the fusion of information from indoor and outdoor photos results in a performance boost of up to 5\% in Macro F1-score.

\end{abstract}

\section{INTRODUCTION}

Over the last few years, significant progress has been made in the field of automatic real estate appraisal.
While earlier models have exclusively utilized textual and categorical input data such as the number of rooms or the floor area \cite{cain_real_1995,park_using_2015,yeh_building_2018} to predict building attributes, recent research has demonstrated that the inclusion of visual information from building photographs can be beneficial \cite{poursaeed_vision-based_2018, kostic_what_2020, you_image-based_2017}. 
Examples include sophisticated price estimation models \cite{poursaeed_vision-based_2018}, machine learning methods for predicting building heating energy demand \cite{despotovic_prediction_2019}, but also
the analysis methods for architectural style \cite{doersch2012makes}.
A prerequisite for the development of efficient machine learning models in the domain of automatic real estate valuation is the availability of a sufficiently large and well-annotated dataset.
In practice, obtaining enough data is usually not an issue, but the corresponding annotations are often incomplete or include varying annotation categories/schemes when obtained from different sources.
This calls for new automated methods to fill such annotation gaps and missing data.

\begin{figure}[thpb]
      \centering
      \includegraphics[scale=0.5]{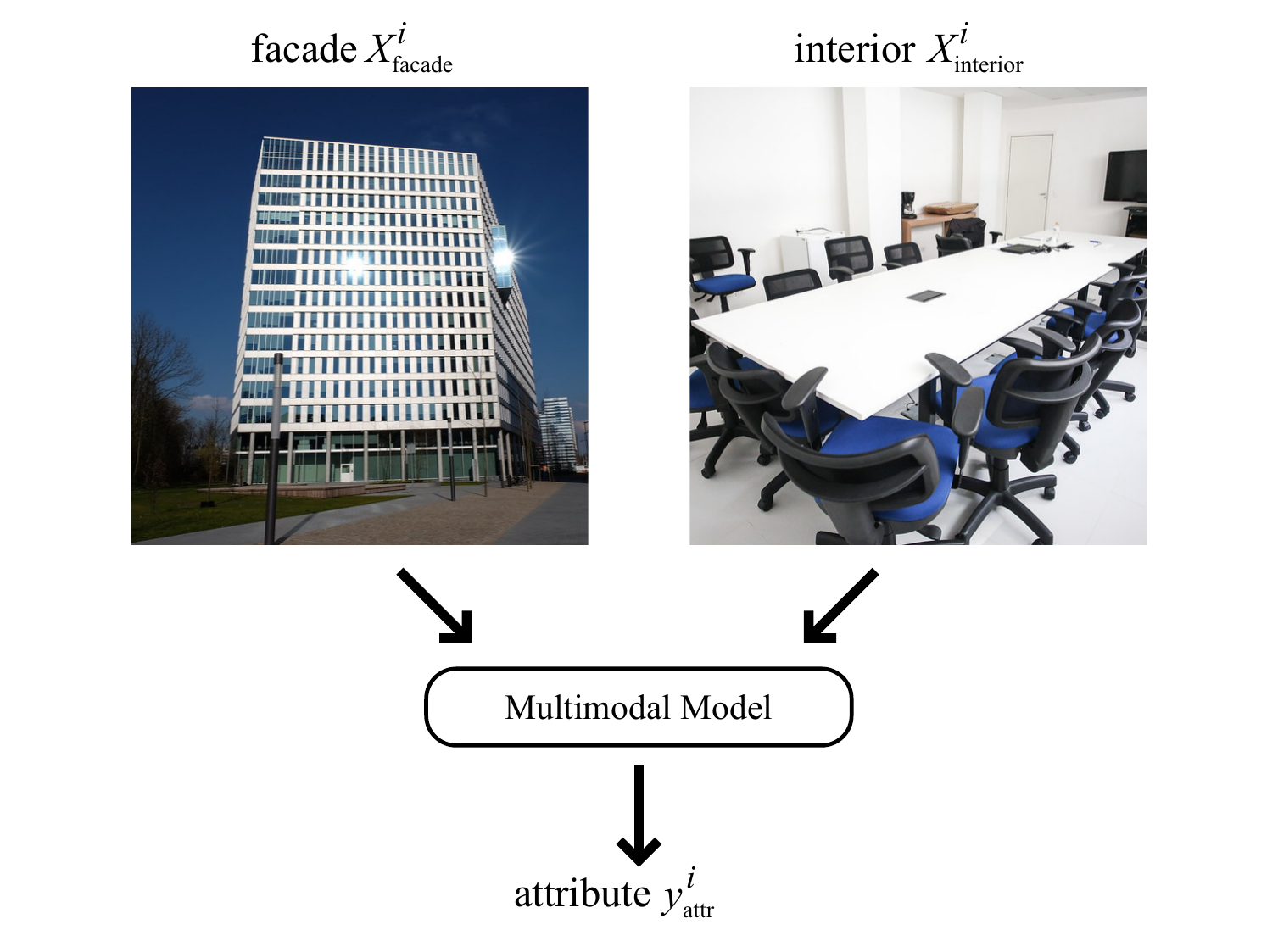}
      \caption{Concept of multimodal learning with two visual modalities.}
      \label{fig:overview}
  \end{figure}

In this work$^{3}$, we leverage the information contained in real estate images to  predict high-level real estate attributes and thereby show a novel way to  fill  missing data in real estate databases. Examples for such attributes that we examine are e.g. the type of commercial use of an object (e.g. ``industrial'', ``hospitality'', ``retail'' or ``office'') or the general type of a building, i.e., whether it is a commercial building or a residential building.
Specifically, we use pairs of facade and interior photos of real estate objects as input which we refer to as two different visual input modalities in the following. This means that the input to our method is a pair of indoor and outdoor images, see also \Cref{fig:overview}. 
The facade and interior embody separate visual aspects of the same property and contain complementary clues for estimating a particular attribute.
Consider the photo pair of \Cref{fig:overview} as an example for the task of differentiating between commercial and residential real estate objects.
The large window fronts of the facade image serve as an indicator that this object may be a commercial office building.
Even stronger hints are provided by the many office chairs in the interior image.
This example illustrates that for each of the two visual input modalities, different types of information need to be extracted and fused to successfully predict a particular attribute.
To evaluate how this can be best achieved, in this work we implement and evaluate three multimodal architectures representing  different fusion approaches with different fusion levels.
In addition, interior and facade photos are not always both available for each real estate object. 
We therefore analyze how robust our proposed models are to missing modalities and whether using additional incomplete samples in the training set can improve prediction accuracy.

\section{Related Work}

In this section we first provide an overview of computer vision methods for real estate analysis and then review related work on multimodal image classification and prediction from missing data/modalities.

\subsection{Real Estate Image Analysis}
An early approach on multimodal learning for real estate analysis, which also utilizes visual information, was proposed by Ahmed et al. \cite{ahmed_house_2016}.
To leverage the image information of a building, the authors extracted SURF features \cite{bay_surf_2006} from different room types and trained a neural network to predict the price from both visual and textual features.
In another work by Kostic et al. \cite{kostic_what_2020}, image entropy, level of greenness, and features extracted from a CNN pretrained on ImageNet \cite{deng_imagenet_2009} were used for price prediction.
A method for estimating the age of a building from its visual appearance was introduced by Zeppelzauer et al. \cite{zeppelzauer_automatic_2018} where the authors extracted patches of interest via SIFT features \cite{lowe_distinctive_2004} and gave them as input to a neural network that predicts the building age through decision fusion. 
This method was extended in Despotovic et al. \cite{despotovic_prediction_2019} for predicting the heating demand of a building.
A model based on long-short-term-memory (LSTM) networks was developed by You et al. \cite{you_image-based_2017}. 
To achieve a robust estimate of a property’s value, the LSTM network was also provided with photos from the neighborhood of the building.
Bin et al. \cite{bin_attention-based_2019} took advantage of attention modules \cite{vaswani_attention_2017} and fused information from both textual data and satellite images in order to automatically predict property prices in Los Angeles.
Using Crowdsourcing, Poursaeed et al. \cite{poursaeed_vision-based_2018} built a dataset with luxury scores for different room types.
Subsequently, a CNN network was trained to predict the luxury score of each room and merge it with textual data to predict the property price. 
A comprehensive overview of the emerging trend of image analysis in the real estate domain has recently been provided by Koch et al. \cite{koch_real_2019}.

\subsection{Multimodal Learning}
An important architectural design choice in multimodal learning is where to fuse the information from different input modalities.
\textit{Early fusion} models combine all modalities at the input level, which can be achieved by concatenating raw data or preprocessed input features \cite{kachele_multimodal_2015,gonzalez_multiview_2015}.
Limitations for this type of models can arise from differing dimensionalities and sampling rates of the input modalities \cite{ramachandram_deep_2017}.
Another option is to fuse modalities at the decision level of the model \cite{glodek_multiple_2011,liu_recognizing_2018,nojavanasghari_deep_2016}, which is usually called \textit{late fusion}.
In this case, a separate classifier is used for each modality, and the overall model prediction can be computed by using e.g. the maximum or average of the predictions or by stacking a meta-classifier on top.
When the information of modalities is merged throughout the model, it is referred to as \textit{intermediate fusion}.
This type of fusion can be achieved in a variety of ways.
Wang et al. \cite{wang_deep_2020} proposed a strategy for handling pairs of corresponding RGB images and depth maps.
Based on the batch normalization activation levels of the model’s intermediate layers, feature map channels are exchanged between both modalities to replace irrelevant information.
The work of Nagrani et al. \cite{nagrani_attention_2021} has shown that Visual Transformers \cite{dosovitskiy_image_2021} can be successfully applied to a multimodal problem.
To exchange cross-modal information in the model they used attention bottlenecks.
In our study we apply the ideas of Joze et al. \cite{joze_mmtm_2020} for one of our three network variants.
The authors used so-called multi modal transfer modules (MMTM) between  modality-specific CNN streams.
These modules help to recalibrate the magnitude of channel-wise features in each stream, which will be described in more detail in section \ref{sec:approach}.

\subsection{Missing Modalities}
Sun et al. \cite{sun_semi-supervised_2021} proposed an image translation method that can compensate for the absence of single modalities.
They implemented an encoder-decoder architecture for each modality and arranged them in a cyclical structure during training so that one image modality can always be reconstructed from the encoded information of another modality.
In a similar approach, Tran et al. \cite{tran_missing_2017} developed a cascading network of residual autoencoders for the task of predicting missing modalities.
Choi et al. \cite{choi_embracenet_2019} used subnetworks for each modality, each yielding a feature vector of the same dimension. 
Then, a random sampling process is applied which takes sparse features from each modality and combines them, improving the ability of the network to  compensate for missing information.
In our work, the ability of our models to handle missing data is not achieved through the network architecture design, but through data augmentation.

\section{Approach}
\label{sec:approach}
The main goal of our work is to develop a network architecture that can perform the following functions.
\begin{enumerate}
    \item When provided with an input pair of both a photo of the building facade $X^{i}_{\mathrm{facade}}$ and from the interior $X^{i}_{\mathrm{interior}}$ of the same real estate object $i$, it should be able to predict the correct class $y^{i}_{\mathrm{attr}}$ of a given category (see \Cref{fig:overview}).
    \item The model should be capable of dealing with missing modalities, which in this instance refers to either an absent indoor $X^{i}_{\mathrm{interior}}$ or facade photo $X^{i}_{\mathrm{facade}}$.
\end{enumerate}
In our method, we handle a missing modality by representing the missing $X^{i}_{\mathrm{interior}}$ or $X^{i}_{\mathrm{facade}}$ as a black image with all RGB values set to zero.
We further investigate how different fusion strategies perform in this scenario.
To this end, we implement three model architectures, each representing a different fusion archetype.
A full description of these architectures can be found in Section \ref{subsec:arch}. The high level attributes which we investigate are the commercial type, residential type and object type of a property.
More details on these attributes can be found in \ref{sec:datasets}
To evaluate our approach, we formulate the following five research questions (RQs), which we will answer in Section~\ref{sec:experimentsResults}.

\begin{itemize}
 \item RQ1: What predictive performance can be achieved for different high-level real estate attributes? 
 \item RQ2: How efficient is the fusion of modalities compared to using only single modalities during training? 
 \item RQ3: What is the best fusion strategy to merge the information of the two input modalities?
 \item RQ4: Are networks trained on complete pairs of photos still capable of correctly predicting missing modality samples?
 \item RQ5: Does the addition of incomplete data in the training set lead to better test accuracy?
\end{itemize}

\subsection{Multimodal Network Architectures}
\label{subsec:arch}

\begin{figure}[thpb]
      \centering
      \includegraphics[scale=0.45]{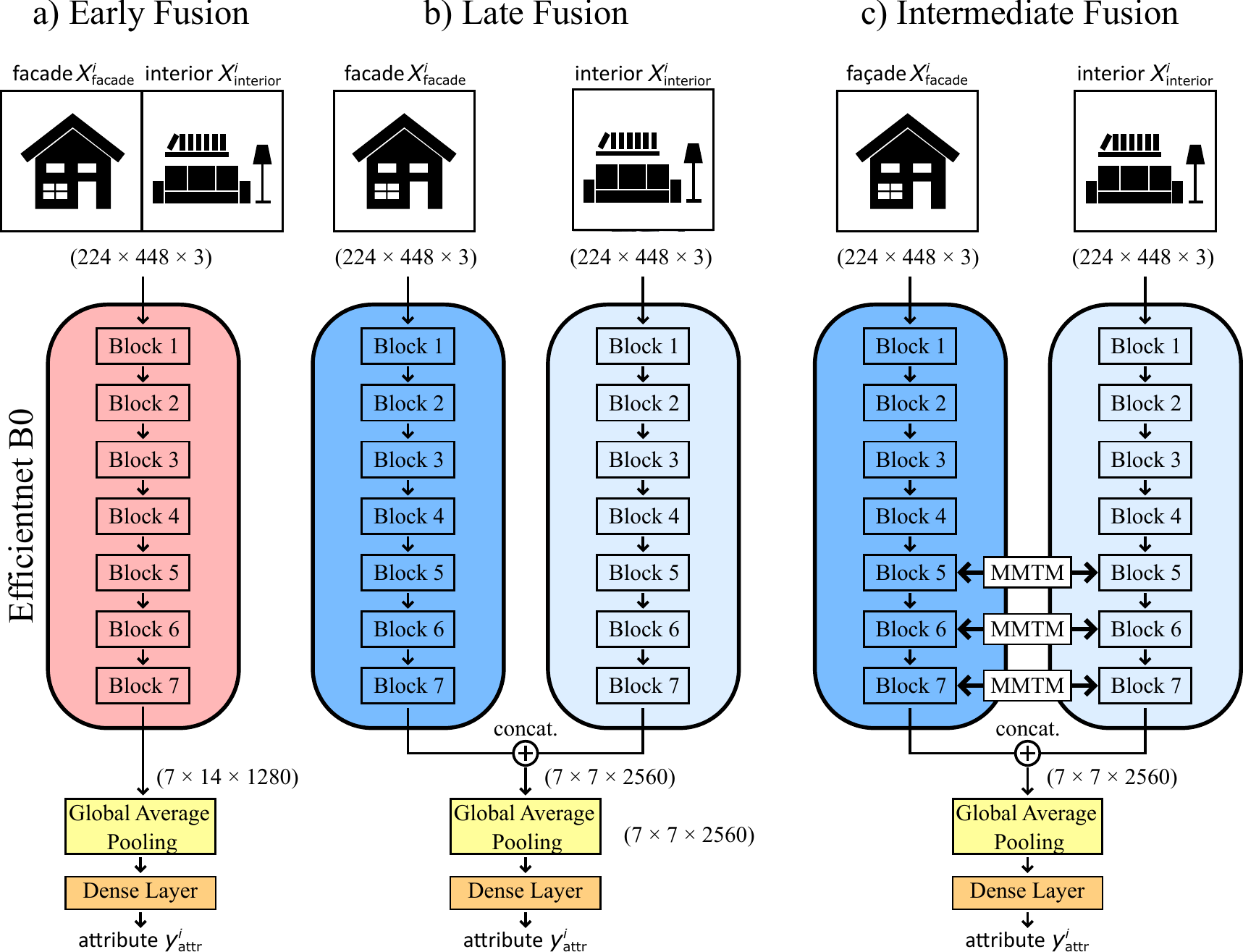}
      \caption{Overview of the developed network architectures.}
      \label{fig:architectures}
  \end{figure}

The key to multimodal classification lies in the effective fusion of information from different modalities.
Therefore, in this work we evaluate the performance of three model architectures that follow different fusion strategies.
For all three architectures EfficientNet B0 \cite{tan_efficientnet_2019} pretrained on ImageNet \cite{deng_imagenet_2009} is chosen as the backbone architecture to achieve strong classification performance and to allow a fair comparison between all architectures.
The three multimodal architecture variants are illustrated in \Cref{fig:architectures} and described in the following.

\begin{figure}[thpb]
      \centering
      \includegraphics[scale=0.45]{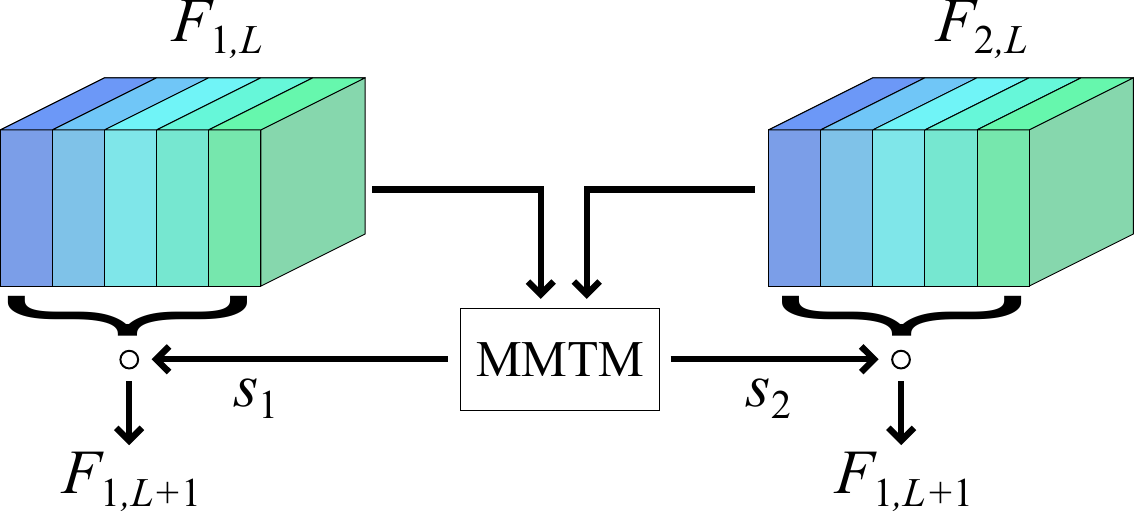}
      \caption{Concept of multimodal transfer modules (MMTM). $F_{1,L}$, $F_{2,L}$ indicate feature maps of both network streams at layer $L$. $s_{1}$, $s_{2}$ are the generated gating signals. }
      \label{fig:mmtm}
  \end{figure}

\textbf{Early Fusion:}
The network architecture in \Cref{fig:architectures} a) represents the concept of early fusion.
Both $X^{i}_{\mathrm{facade}}$ and $X^{i}_{\mathrm{interior}}$ of every input pair, each of size (224 x 224 x 3) are horizontally concatenated at the beginning to produce a single input image of size (224 x 448 x 3).
The concatenated samples are then fed to the EfficientNet B0 backbone, whose output is a featuremap of size (7 x 14 x 1280).
This layer is followed by a global average pooling and a dense layer with softmax activation to output the classification scores.
\\\\
\textbf{Late Fusion:}
Here, instead of concatenating the input images at the beginning, both image modalities are processed in separate subnetworks and are fused at a later stage (\Cref{fig:architectures} b)).
Therefore, two separate EfficientNet B0 sub-networks are utilized, which accept input images of size (224 x 224 x 3).
In the fusion stage, the two (7 x 7 x 1280) output feature maps are concatenated along the channel dimension and are again processed through a global average pooling layer and a dense layer.
\\\\
\textbf{Intermediate Fusion:}
The third architecture in \Cref{fig:architectures} c) is an extension of the previous one with multimodal transfer modules (MMTM) introduced by Joze et al. \cite{joze_mmtm_2020}.
The concept behind multimodal transfer module blocks is illustrated in  \Cref{fig:mmtm}.
An MMTM block accepts two feature maps $F_{1,L}$, $F_{2,L}$ from the same Layer $L$ of the two network streams 1 and 2. Within the MMTM block, the information from both feature maps then gets merged through global average pooling and dense layers to generate two gating signals $s_1$ and $s_2$.
Both gating signals are used to reweight the importance of each featuremap channel of $F_{1,L}$ and $F_{2,L}$.
For more details the interested reader can refer to \cite{joze_mmtm_2020}.
We use three MMTM blocks, which connect the outputs of the first excitation layers of stages 5, 6 and 7 of EfficientNet B0 \cite{tan_efficientnet_2019}.


\section{Experimental and Results}
\label{sec:experimentsResults}
In this section, we first provide an overview of the datasets and use cases that serve for the evaluation of our approach. 
Furthermore, we provide the training details, including the used hyperparameters and the evaluation metrics.

\subsection{Datasets and Use Cases}
\label{sec:datasets}


\begin{table*}[ht]
\centering
\caption{DATASETS FOR THE THREE INVESTIGATED USE CASES}
\label{tab:datasets}
\begin{tabular}{|c|c|c|c|c|c|c|c|c|c|c|}
\cline{1-5} \cline{7-8} \cline{10-11}
\multirow{2}{*}{dataset split} & \multicolumn{4}{c|}{UC1: Commercial type} &  & \multicolumn{2}{c|}{UC2: Residential type} &  & \multicolumn{2}{c|}{UC3: Object type} \\ \cline{2-5} \cline{7-8} \cline{10-11} 
 & \multicolumn{1}{c|}{industry} & \multicolumn{1}{c|}{hospitality sector} & \multicolumn{1}{c|}{retail} & offices &  & \multicolumn{1}{c|}{apartment} & house &  & \multicolumn{1}{c|}{commercial} & residential \\ \cline{1-5} \cline{7-8} \cline{10-11} 
Train & 25 (+30) & 30 (+20) & 75 (+100) & 100 (+50) &  & 300 (+250) & 300 (+250) &  & 230 (+200) & 600 (+500) \\ \cline{1-1}
Val & 12 (+14) & 15 (+10) & 37 (+40) & 47 (+20) &  & 50 (+50) & 50 (+50) &  & 111 (+84) & 100 (+100) \\ \cline{1-1}
Test & 14 & 17 & 43 & 50 &  & 667 & 177 &  & 124 & 844 \\ \cline{1-5} \cline{7-8} \cline{10-11}

\end{tabular}

\end{table*}

We evaluate our approach with three different sets of real estate categories and therefore compile the following datasets with respective class labels, making up three different use cases (UC) for evaluation:
\begin{itemize}
 \item UC1 - Commercial type: classes: industrial, hospitality sector, retail, office
 \item UC2 - Residential type: classes: apartment, house
 \item UC3 - Object type: classes: commercial, residential
\end{itemize}
Each of the respective datasets consists of pairs of facade and indoor photos taken from real estate objects in Austria with corresponding class labels. Often, there are several interior and exterior photos per real estate object.
We handle this case by creating multiple unique samples for each real estate object. For example, if six interior and three exterior photos are available for an ``office'' class commercial object, we create three interior-exterior pair samples of ground-truth class ``office'' by selecting three random interior photos and assigning one outdoor photo to each. Regardless of whether there are multiple pairs of photos per real estate object, all generated samples are assigned the ground truth class of the associated real estate property.
\\\\
An overview of these datasets, classes and their partitioning into training, validation and test set can be found in \Cref{tab:datasets}.
In our experiments we also want to investigate whether training with additional incomplete data, meaning either indoor $X^{i}_{\mathrm{interior}}$ or facade image $X^{i}_{\mathrm{facade}}$ is missing, can lead to an improvement in prediction accuracy.
Therefore, we optionally add incomplete samples to the datasets, where the respective missing visual modality is replaced by a black image.
The amount of additional incomplete samples is indicated by the values in parentheses in \Cref{tab:datasets}.
When only complete samples are used during training, we refer to the dataset as ``\textit{complete}'' and when additional missing samples are added we denote it as ``\textit{complete} + \textit{missing}''.
To avoid bias in favor of one modality, the number of samples with missing facades and missing interior in the ``Missing'' dataset is kept equal.

\subsection{Training Procedure and Parameters}

All experiments are conducted with the following hyperparameters.
Training is performed for a total of 200 epochs with a batch size of 16 and a learning rate of 0.0001 using the Adam optimizer.
As a loss function, categorical cross entropy is used.
After each epoch, the updated network weights are only saved if the validation loss decreases.
To prevent overfitting, we also apply several data augmentation operations including image flipping, rotation, zoom, shear and brightness correction.
If an incomplete sample is fed to the network we replace the missing modality with a black image.

\subsection{Evaluation Metric}

Since we have a varying amount of data available for each class, our test sets also have different numbers of samples.
In our evaluation we nevertheless want to give equal importance to each class and therefore use the Macro F1-score metric, which is defined as follows:

\begin{equation}
\mathrm{Macro \,\,\, F1\text{-}score} = \frac{1}{N} \sum_{i=1}^{N} \mathrm{F1\text{-}score}_{i},
\end{equation}

where $N$ is the number of classes and $i$ represents the class label.





\subsection{Experiments}

In the following, we provide an overview of our experiments.
We run experiments for variations of different use cases, modality configurations and multimodal architectures (independent variables). Details on each variable are provided below.
\\\\
\textbf{Use Cases:} Each experiment is conducted on all three use cases, where each has its corresponding dataset (see \Cref{tab:datasets}).
\\\\
\textbf{Modality Configuration:} We further want to evaluate whether a multimodal learning approach leads to better results than using only single modality data for training, which is why we also analyze four different modality configurations. The first is the default \textit{complete} configuration, where all data consists of full pairs of interior and facade photos. From this we generate two additional single modality configurations. Specifically, for \textit{facade only} we modify the \textit{complete} configuration by setting all interior photos to black and do the opposite for \textit{interior only}.
Finally, we generate a fourth \textit{complete} + \textit{missing} configuration, in which extra missing modality samples are added to the \textit{complete} configuration (compare \Cref{tab:datasets}).
\\\\
\textbf{Multimodal Architecture:} We conduct each experiment with all three multimodal network architectures (early fusion, late fusion and intermediate fusion, see \Cref{fig:architectures}).
\\\\
In total this amounts to 36 different experiment configurations (3 use cases, 4 dataset configurations, 3 network architectures).
In addition, we repeat every training process three times for each experiment to capture the variations of results originating from different random initializations of the network weights.

\begin{table*}[]
\caption{Macro F1-scores averaged over three training runs and in parentheses the respective standard deviations. RB indicates the random baseline.}
\label{tab:results}
\resizebox{1\textwidth}{!}{
\begin{tabular}{|c|c|ccc|ccc|ccc|}
\hline
\multirow{3}{*}{\makecell{Modality \\ Configuration}} & \multirow{3}{*}{\makecell{Multimodal \\ Architecture}} & \multicolumn{3}{c|}{\multirow{2}{*}{\makecell{UC1: Commercial type \\ (RB = 25\%)} }} & \multicolumn{3}{c|}{\multirow{2}{*}{\makecell{UC2: Residential type \\ (RB = 50\%)}}} & \multicolumn{3}{c|}{\multirow{2}{*}{\makecell{UC3: Object type \\ (RB = 50\%)}}} \\
 &  & \multicolumn{3}{c|}{} & \multicolumn{3}{c|}{} & \multicolumn{3}{c|}{} \\ \cline{3-11} 
 &  & \multicolumn{1}{c|}{test\_c} & \multicolumn{1}{c|}{test\_f} & test\_i & \multicolumn{1}{c|}{test\_c} & \multicolumn{1}{c|}{test\_f} & test\_i & \multicolumn{1}{c|}{test\_c} & \multicolumn{1}{c|}{test\_f} & test\_i \\ \hline
\multirow{3}{*}{\textit{complete}} & early & 0.54 (0.04) & 0.37 (0.05) & 0.42 (0.03) & \textbf{0.78 (0.01)} & 0.76 (0.01) & 0.58 (0.01) & 0.76 (0.04) & 0.71 (0.03) & 0.70 (0.03) \\ \cline{2-2}
 & late & 0.54 (0.06) & 0.36 (0.05) & 0.42 (0.03) & 0.76 (0.01) & 0.76 (0.01) & 0.60 (0.01) & 0.77 (0.02) & \textbf{0.72 (0.01)} & 0.71 (0.01) \\ \cline{2-2}
 & intermediate & 0.56 (0.05) & 0.41 (0.04) & 0.44 (0.02) & 0.77 (0.01) & 0.76 (0.01) & 0.61 (0.01) & 0.77 (0.02) & \textbf{0.72 (0.01)} & 0.72 (0.01) \\ \hline
\multirow{3}{*}{\textit{facade only}} & early &  & \textbf{0.52 (0.03)} &  &  & 0.72 (0.01) &  &  & 0.70 (0.02) &  \\ \cline{2-2}
 & late &  & 0.40 (0.06) &  &  & 0.75 (0.01) &  &  & 0.66 (0.03) &  \\ \cline{2-2}
 & intermediate &  & 0.41 (0.03) &  &  & \textbf{0.77 (0.01)} &  &  & 0.70 (0.02) &  \\ \hline
\multirow{3}{*}{\textit{interior only}} & early &  &  & 0.41 (0.04) &  &  & 0.60 (0.01) &  &  & 0.69 (0.01) \\ \cline{2-2}
 & late &  &  & 0.44 (0.02) &  &  & 0.61 (0.01) &  &  & 0.72 (0.01) \\ \cline{2-2}
 & intermediate &  &  & 0.42 (0.05) &  &  & \textbf{0.62 (0.01)} &  &  & 0.67 (0.06) \\ \hline
\multirow{3}{*}{\textit{complete}+\textit{missing}} & early & 0.57 (0.03) & 0.40 (0.03) & 0.45 (0.03) & 0.75 (0.01) & 0.72 (0.02) & 0.57 (0.01) & 0.77 (0.01) & 0.71 (0.01) & 0.71 (0.03) \\ \cline{2-2}
 & late & \textbf{0.62 (0.03)} & 0.42 (0.08) & \textbf{0.49 (0.02)} & 0.75 (0.01) & 0.73 (0.01) & 0.58 (0.05) & 0.79 (0.02) & \textbf{0.72 (0.02)} & 0.70 (0.02) \\ \cline{2-2}
 & intermediate & \textbf{0.62 (0.03)} & 0.42 (0.06) & 0.48 (0.05) & 0.76 (0.01) & 0.74 (0.01) & 0.55 (0.03) & \textbf{0.81 (0.01)} & \textbf{0.72 (0.01)} & \textbf{0.75 (0.00)} \\ \hline
\end{tabular}

}

\end{table*}

\section{Results}

In the following, we present our experimental results and answer the posed research questions from Section \ref{sec:approach}. 
The results of all our 36 experiments can be found in \Cref{tab:results}.
The presented values are Macro F1-scores for the respective test sets, which are additionally averaged over all three training runs.
The value inside the parentheses is the standard deviation over all three training repetitions.
To evaluate the performance when a network receives samples with missing modality, the same test set is used in three alterations.
\emph{test\textunderscore c} refers to the test set with complete pairs (no missing data). 
\emph{test\textunderscore f} and \emph{test\textunderscore i} refer to the same test set, but here only one modality, facade or interior, is used at a time, while the other one is blackened to simulate missing data in the test sets. For an overview of the split for each modality configuration refer to section \ref{sec:datasets}.
\\\\
With respect to research question 1 (RQ1), \Cref{tab:results} shows that the prediction scores differ greatly between the different use cases. 
While the best Macro F1-score for UC1 (Commercial type) is 0.62, the highest prediction value for UC2 (Residential type) amounts to 0.78.
In the UC3 (Object type) setting, the Macro F1-score reaches 0.81.
However, it should be noted that the random baseline (RB) of 50\% for UC2 and UC3 is already much bigger than the respective 25\% of UC1. Nevertheless, a large margin over the random baseline is achieved for all three use cases. 
\\\\
With research question 2 (RQ2) we wanted to discern whether multimodal learning on both visual modalities is superior to training on individual modalities.
For all use cases, \textit{complete} yields better results than \textit{facade only} and \textit{interior only}. 
There is an increase of 4\% of the score for UC1 compared to the best result for the single modality configurations. 
For UC3, the improvement is 5\%. 
Only for UC2 the performances are almost equal.
The reason for the high score for residential properties is probably due to the strong difference in the appearance of facades of apartment buildings and houses, which is also reflected in the similarly high score of the \textit{facade only} configuration.
Overall, we can see that training on both modalities provides clear advantages over using only one modality.
\\\\
Regarding research question 3 (RQ3: which architecture is best suited for multimodal fusion?) we do not reach a clear conclusion.
In almost all cases Macro F1-score differences are within 1\% or 2\%, which does not allow for declaring a clear winner when considering the standard deviations across the three runs. 
One possible explanation for why the early fusion architecture produces similar results compared to the others, is the fact that both visual modalities concatenated at the input level are RGB images.
Hence, the network does not have to deal with information of different dimensionality and domains in its initial layers.
It can therefore focus on learning to extract the same low-level features (e.g. edges), which are representative for both input modalities. 
To summarize the answer to RQ3, we find no significant performance differences between using early, late and intermediate fusion strategies in the evaluated use cases.
\\\\
Concerning research question 4 (RQ4: generalizability and robustness to missing data) we compare the Macro F1-scores of the \textit{complete} configuration for \emph{test\textunderscore f} and \emph{test\textunderscore i} with that of the training configurations \textit{interior only} and \textit{facade only}.
Despite the fact that the corresponding networks of \textit{complete} have never been exposed to missing modalities and have only been trained on complete samples they still provide comparable prediction scores for  \emph{test\textunderscore f} and \emph{test\textunderscore i}.
Overall the results show that our multimodal network architectures are capable of handling incomplete input data.
\\\\
Investigating research question 5 (RQ5) shows that adding additional data with missing modalities leads to better results for two of three use cases. 
In case of UC1, the increase in Macro F1-score from training on \textit{complete} to \textit{complete} + \textit{missing} is the largest with almost 6\%.
For UC2, scores are at the same level, whereas for UC3 performance increases by 6\%.
These results show that the proposed multimodal network architectures can take benefit of the information contained in the additional incomplete training samples.

\begin{figure*}[thpb]
      \centering
      \includegraphics[scale=0.23]{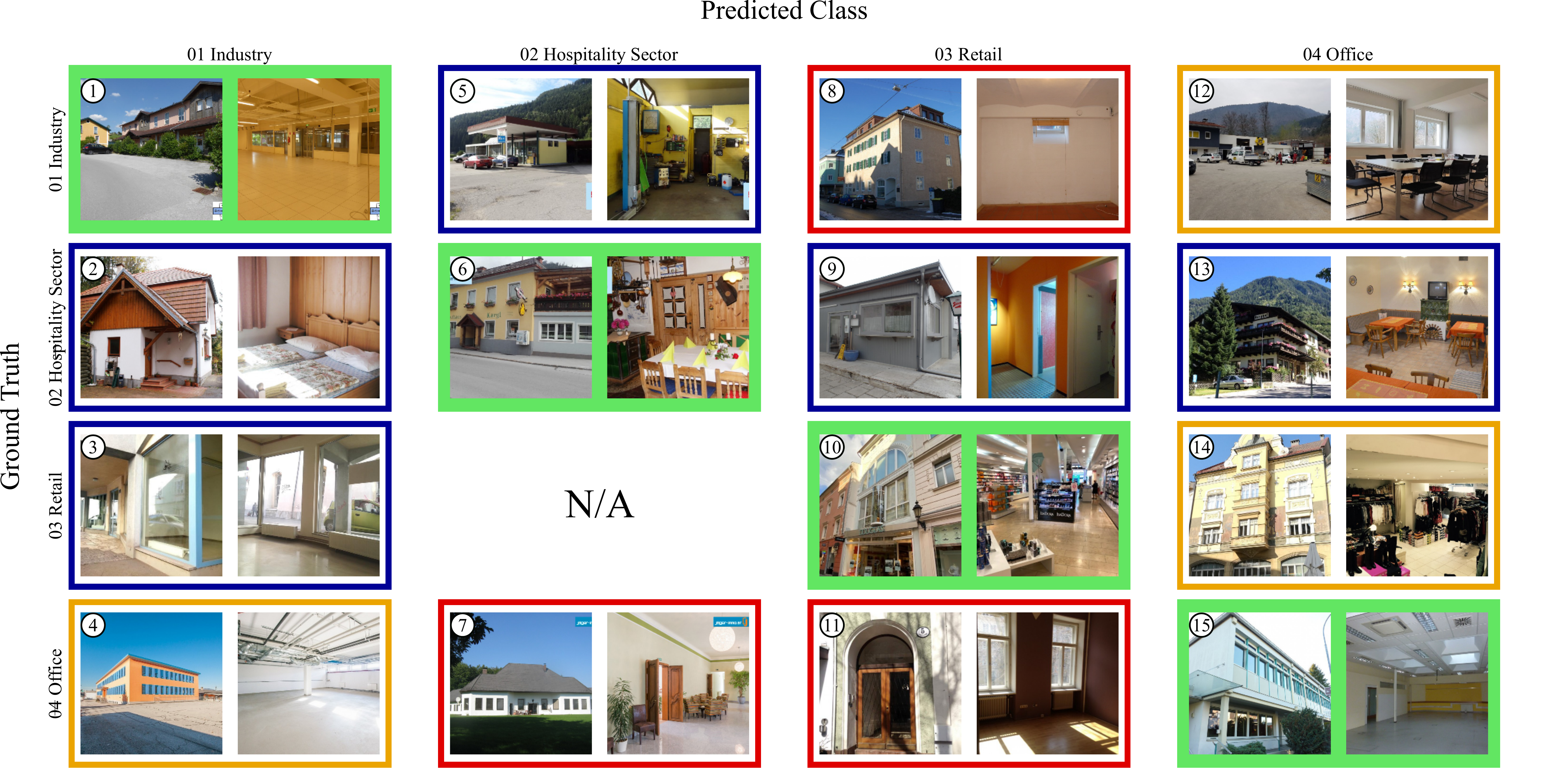}
      \caption{Confusion matrix with exemplary predicted images from the testset. A Green Background indicates true positive samples. Colored borders indicate different failure types (blue: unused clues, orange: conflicting clues, red: missing clues). Photos taken from justimmo\protect\footnotemark[4] .}
      \label{fig:confusion}
  \end{figure*}

\section{Qualitative Results}

To further investigate especially the limitations of our approach, we qualitatively analyzed the results. 
During our experiments, we found that pairs of images that were incorrectly predicted by our networks can be systematically grouped into three main failure types.
In this section we want to showcase these failure types using exemplary pairs of photos from our test set and their corresponding predicted labels.
For this purpose, we take UC1 (commercial types) and the predictions from the late multimodal architecture for the \textit{complete} + \textit{missing} modality configuration because it represents one of the most robust combinations.
The selected pairs of indoor and facade photos are shown in \Cref{fig:confusion}.
All pairs are placed in a confusion matrix-like layout, with true positive samples indicated by a green background (diagonal samples).
The three failure types are represented by different border colors for the off-diagonal entries.
\\\\
\textbf{Unused Clues (blue):} This failure type includes samples whose class can be easily recognized by the human observer, but which was not predicted correctly by the network.
For example, pair 2) shows two beds in the interior, which is a clear indication for a hospitality object.
In addition, in pair 3), the depicted retail property was also misclassified as an industrial building despite having a visible storefront.
One explanation for the failed detection in this case could be that in our dataset many industrial buildings have a gray colored floor similar to the one in this pair.
In image 9), a lamp post with a brewery logo can be seen, which is a subtle hint for a restaurant that a human observer can understand but was not detected by the network.
We hypothesize that this failure type can be mitigated by increasing the total amount of training data available. 
This way, the network receives more samples from which it can learn relevant patterns.
\\\\
\textbf{Conflicting Clues (orange):} Some of the samples shown have visual modalities that contain conflicting information.
The pair 4) shows photos of an office building with a corresponding looking facade. 
However, the interior photo depicts a large hall that could also be found in a typical industrial building. 
The opposite case for an actual industry building can be found in pair 12).
Here, the interior photo displays a conference room suggestive of an office building, whereas the exterior resembles an industry building.
Pair 14) is a clothing store, which can be recognized by the interior photo.
The facade, on the other hand, has nothing in common with typical storefronts. 
To reduce this failure type, increasing the size of the dataset alone may not be sufficient.
In practice, there are often more than two photos available for a given property, all of which could be used in a single model to counteract conflicting modalities. Furthermore, to mitigate such cases, it will be important to assess the representativeness of an image for the target class, i.e., to give less characteristic and speaking images less weight.
\\\\
\textbf{Missing Clues (red):} The last failure type contains samples that lack any useful clues for classification.
In pair 7) a real estate object with an unusual appearance for an office building is shown, which represents a difficult task for our network.
Example 11) contains a pair of photos with little useful information.
The outdoor photo is a close-up of the door, that gives no hints about the rest of the facade, and the interior photo is a shot of an empty room in suboptimal lighting conditions.
A similar issue is present in pair 8). 
The facade is ambiguous and the room is also empty and lacks information.
With respect to this type of failure, the use of additional input photos per property could also be beneficial.
In practice, however, we expect that for a certain percentage of real estate objects accurate predictions will fail due to  ambiguous or inexpressive pictures. In such cases the incorporation of additional data modalities, e.g. textual descriptions and categorical data can help.

\section{Conclusion}

In this paper, we demonstrated the effectiveness and feasibility of using visual data  for the prediction of high-level real estate attributes. 
We leveraged two complementary visual modalities, compared different multimodal fusion strategies and evaluated our approach in three different use cases.
Our experiments show that networks trained on both visual modalities (facade and interior) yield better results than networks utilizing only one modality.  
Furthermore, we could show that our multimodal network architectures  provide robust predictions for input samples, which lack one of the two input modalities and that additional training data -- even when it is incomplete -- can improve the robustness of the models.
In future, we plan to extend the proposed multimodal architectures to accept an arbitrary number of input images showing different perspectives of a real estate object.
\footnotetext[4]{www.justimmo.at}

\addtolength{\textheight}{0cm} 







{\small
\bibliographystyle{IEEEtranS}
\bibliography{paper}

\begin{thebibliography}{10}
\providecommand{\url}[1]{#1}
\csname url@rmstyle\endcsname
\providecommand{\newblock}{\relax}
\providecommand{\bibinfo}[2]{#2}
\providecommand\BIBentrySTDinterwordspacing{\spaceskip=0pt\relax}
\providecommand\BIBentryALTinterwordstretchfactor{4}
\providecommand\BIBentryALTinterwordspacing{\spaceskip=\fontdimen2\font plus
\BIBentryALTinterwordstretchfactor\fontdimen3\font minus
  \fontdimen4\font\relax}
\providecommand\BIBforeignlanguage[2]{{%
\expandafter\ifx\csname l@#1\endcsname\relax
\typeout{** WARNING: IEEEtran.bst: No hyphenation pattern has been}%
\typeout{** loaded for the language `#1'. Using the pattern for}%
\typeout{** the default language instead.}%
\else
\language=\csname l@#1\endcsname
\fi
#2}}

\bibitem{ahmed_house_2016}
\BIBentryALTinterwordspacing
E.~Ahmed and M.~Moustafa, ``House price estimation from visual and textual
  features,'' \emph{arXiv:1609.08399 [cs]}, Sept. 2016, arXiv: 1609.08399.
  [Online]. Available: \url{http://arxiv.org/abs/1609.08399}
\BIBentrySTDinterwordspacing

\bibitem{bay_surf_2006}
H.~Bay, T.~Tuytelaars, and L.~Van~Gool, ``\BIBforeignlanguage{en}{{SURF}:
  {Speeded} {Up} {Robust} {Features}},'' in
  \emph{\BIBforeignlanguage{en}{Computer {Vision} – {ECCV} 2006}}, ser.
  Lecture {Notes} in {Computer} {Science}, A.~Leonardis, H.~Bischof, and
  A.~Pinz, Eds.\hskip 1em plus 0.5em minus 0.4em\relax Berlin, Heidelberg:
  Springer, 2006, pp. 404--417.

\bibitem{bin_attention-based_2019}
\BIBentryALTinterwordspacing
J.~Bin, B.~Gardiner, Z.~Liu, and E.~Li,
  ``\BIBforeignlanguage{en}{Attention-based multi-modal fusion for improved
  real estate appraisal: a case study in {Los} {Angeles}},''
  \emph{\BIBforeignlanguage{en}{Multimedia Tools and Applications}}, vol.~78,
  no.~22, pp. 31\,163--31\,184, Nov. 2019. [Online]. Available:
  \url{https://doi.org/10.1007/s11042-019-07895-5}
\BIBentrySTDinterwordspacing

\bibitem{cain_real_1995}
\BIBentryALTinterwordspacing
M.~Cain and C.~Janssen, ``\BIBforeignlanguage{en}{Real estate price prediction
  under asymmetric loss},'' \emph{\BIBforeignlanguage{en}{Annals of the
  Institute of Statistical Mathematics}}, vol.~47, no.~3, pp. 401--414, Sept.
  1995. [Online]. Available: \url{https://doi.org/10.1007/BF00773391}
\BIBentrySTDinterwordspacing

\bibitem{choi_embracenet_2019}
J.-H. Choi and J.-S. Lee, ``{EmbraceNet}: {A} robust deep learning architecture
  for multimodal classification,'' \emph{Information Fusion}, vol.~51, pp.
  259--270, 2019, publisher: Elsevier.

\bibitem{deng_imagenet_2009}
J.~Deng, W.~Dong, R.~Socher, L.-J. Li, K.~Li, and L.~Fei-Fei, ``{ImageNet}: {A}
  large-scale hierarchical image database,'' in \emph{2009 {IEEE} {Conference}
  on {Computer} {Vision} and {Pattern} {Recognition}}, June 2009, pp. 248--255,
  iSSN: 1063-6919.

\bibitem{despotovic_prediction_2019}
\BIBentryALTinterwordspacing
M.~Despotovic, D.~Koch, S.~Leiber, M.~Döller, M.~Sakeena, and M.~Zeppelzauer,
  ``\BIBforeignlanguage{en}{Prediction and analysis of heating energy demand
  for detached houses by computer vision},''
  \emph{\BIBforeignlanguage{en}{Energy and Buildings}}, vol. 193, pp. 29--35,
  June 2019. [Online]. Available:
  \url{https://www.sciencedirect.com/science/article/pii/S0378778818336430}
\BIBentrySTDinterwordspacing

\bibitem{doersch2012makes}
C.~Doersch, S.~Singh, A.~Gupta, J.~Sivic, and A.~Efros, ``What makes paris look
  like paris?'' \emph{ACM Transactions on Graphics}, vol.~31, no.~4, 2012.

\bibitem{dosovitskiy_image_2021}
\BIBentryALTinterwordspacing
A.~Dosovitskiy, L.~Beyer, A.~Kolesnikov, D.~Weissenborn, X.~Zhai,
  T.~Unterthiner, M.~Dehghani, M.~Minderer, G.~Heigold, S.~Gelly, J.~Uszkoreit,
  and N.~Houlsby, ``An {Image} is {Worth} 16x16 {Words}: {Transformers} for
  {Image} {Recognition} at {Scale},'' \emph{arXiv:2010.11929 [cs]}, June 2021,
  arXiv: 2010.11929. [Online]. Available: \url{http://arxiv.org/abs/2010.11929}
\BIBentrySTDinterwordspacing

\bibitem{glodek_multiple_2011}
M.~Glodek, S.~Tschechne, G.~Layher, M.~Schels, T.~Brosch, S.~Scherer,
  M.~Kächele, M.~Schmidt, H.~Neumann, G.~Palm, and F.~Schwenker,
  ``\BIBforeignlanguage{en}{Multiple {Classifier} {Systems} for the
  {Classification} of {Audio}-{Visual} {Emotional} {States}},'' in
  \emph{\BIBforeignlanguage{en}{Affective {Computing} and {Intelligent}
  {Interaction}}}, ser. Lecture {Notes} in {Computer} {Science}, S.~D’Mello,
  A.~Graesser, B.~Schuller, and J.-C. Martin, Eds.\hskip 1em plus 0.5em minus
  0.4em\relax Berlin, Heidelberg: Springer, 2011, pp. 359--368.

\bibitem{gonzalez_multiview_2015}
A.~González, G.~Villalonga, J.~Xu, D.~Vázquez, J.~Amores, and A.~M. López,
  ``Multiview random forest of local experts combining {RGB} and {LIDAR} data
  for pedestrian detection,'' in \emph{2015 {IEEE} {Intelligent} {Vehicles}
  {Symposium} ({IV})}, June 2015, pp. 356--361, iSSN: 1931-0587.

\bibitem{joze_mmtm_2020}
H.~R.~V. Joze, A.~Shaban, M.~L. Iuzzolino, and K.~Koishida, ``Mmtm: Multimodal
  transfer module for cnn fusion,'' in \emph{Proceedings of the IEEE/CVF
  Conference on Computer Vision and Pattern Recognition}, 2020, pp.
  13\,289--13\,299.

\bibitem{koch_real_2019}
\BIBentryALTinterwordspacing
D.~Koch, M.~Despotovic, S.~Leiber, M.~Sakeena, M.~Döller, and M.~Zeppelzauer,
  ``Real {Estate} {Image} {Analysis}: {A} {Literature} {Review},''
  \emph{Journal of Real Estate Literature}, vol.~27, no.~2, pp. 269--300, Dec.
  2019, publisher: Routledge \_eprint:
  https://doi.org/10.22300/0927-7544.27.2.269. [Online]. Available:
  \url{https://doi.org/10.22300/0927-7544.27.2.269}
\BIBentrySTDinterwordspacing

\bibitem{kostic_what_2020}
Z.~Kostic and A.~Jevremovic, ``What {Image} {Features} {Boost} {Housing}
  {Market} {Predictions}?'' \emph{IEEE Transactions on Multimedia}, vol.~22,
  no.~7, pp. 1904--1916, July 2020, conference Name: IEEE Transactions on
  Multimedia.

\bibitem{kachele_multimodal_2015}
M.~Kächele, P.~Thiam, M.~Amirian, P.~Werner, S.~Walter, F.~Schwenker, and
  G.~Palm, ``\BIBforeignlanguage{en}{Multimodal {Data} {Fusion} for
  {Person}-{Independent}, {Continuous} {Estimation} of {Pain} {Intensity}},''
  in \emph{\BIBforeignlanguage{en}{Engineering {Applications} of {Neural}
  {Networks}}}, ser. Communications in {Computer} and {Information} {Science},
  L.~Iliadis and C.~Jayne, Eds.\hskip 1em plus 0.5em minus 0.4em\relax Cham:
  Springer International Publishing, 2015, pp. 275--285.

\bibitem{liu_recognizing_2018}
M.~Liu and J.~Yuan, ``Recognizing {Human} {Actions} as the {Evolution} of
  {Pose} {Estimation} {Maps},'' 2018, pp. 1159--1168.

\bibitem{lowe_distinctive_2004}
\BIBentryALTinterwordspacing
D.~G. Lowe, ``\BIBforeignlanguage{en}{Distinctive {Image} {Features} from
  {Scale}-{Invariant} {Keypoints}},''
  \emph{\BIBforeignlanguage{en}{International Journal of Computer Vision}},
  vol.~60, no.~2, pp. 91--110, Nov. 2004. [Online]. Available:
  \url{https://doi.org/10.1023/B:VISI.0000029664.99615.94}
\BIBentrySTDinterwordspacing

\bibitem{nagrani_attention_2021}
\BIBentryALTinterwordspacing
A.~Nagrani, S.~Yang, A.~Arnab, A.~Jansen, C.~Schmid, and C.~Sun, ``Attention
  {Bottlenecks} for {Multimodal} {Fusion},'' \emph{arXiv:2107.00135 [cs]}, June
  2021, arXiv: 2107.00135. [Online]. Available:
  \url{http://arxiv.org/abs/2107.00135}
\BIBentrySTDinterwordspacing

\bibitem{nojavanasghari_deep_2016}
B.~Nojavanasghari, D.~Gopinath, J.~Koushik, T.~Baltrušaitis, and L.-P.
  Morency, ``Deep multimodal fusion for persuasiveness prediction,'' in
  \emph{Proceedings of the 18th {ACM} {International} {Conference} on
  {Multimodal} {Interaction}}, 2016, pp. 284--288.

\bibitem{park_using_2015}
\BIBentryALTinterwordspacing
B.~Park and J.~K. Bae, ``\BIBforeignlanguage{en}{Using machine learning
  algorithms for housing price prediction: {The} case of {Fairfax} {County},
  {Virginia} housing data},'' \emph{\BIBforeignlanguage{en}{Expert Systems with
  Applications}}, vol.~42, no.~6, pp. 2928--2934, Apr. 2015. [Online].
  Available:
  \url{https://www.sciencedirect.com/science/article/pii/S0957417414007325}
\BIBentrySTDinterwordspacing

\bibitem{poursaeed_vision-based_2018}
\BIBentryALTinterwordspacing
O.~Poursaeed, T.~Matera, and S.~Belongie,
  ``\BIBforeignlanguage{en}{Vision-based real estate price estimation},''
  \emph{\BIBforeignlanguage{en}{Machine Vision and Applications}}, vol.~29,
  no.~4, pp. 667--676, May 2018. [Online]. Available:
  \url{https://doi.org/10.1007/s00138-018-0922-2}
\BIBentrySTDinterwordspacing

\bibitem{ramachandram_deep_2017}
D.~Ramachandram and G.~W. Taylor, ``Deep {Multimodal} {Learning}: {A} {Survey}
  on {Recent} {Advances} and {Trends},'' \emph{IEEE Signal Processing
  Magazine}, vol.~34, no.~6, pp. 96--108, Nov. 2017, conference Name: IEEE
  Signal Processing Magazine.

\bibitem{sun_semi-supervised_2021}
W.~Sun, F.~Ma, Y.~Li, S.-L. Huang, S.~Ni, and L.~Zhang, ``Semi-{Supervised}
  {Multimodal} {Image} {Translation} for {Missing} {Modality} {Imputation},''
  in \emph{{ICASSP} 2021 - 2021 {IEEE} {International} {Conference} on
  {Acoustics}, {Speech} and {Signal} {Processing} ({ICASSP})}, June 2021, pp.
  4320--4324, iSSN: 2379-190X.

\bibitem{tan_efficientnet_2019}
\BIBentryALTinterwordspacing
M.~Tan and Q.~Le, ``\BIBforeignlanguage{en}{{EfficientNet}: {Rethinking}
  {Model} {Scaling} for {Convolutional} {Neural} {Networks}},'' in
  \emph{\BIBforeignlanguage{en}{Proceedings of the 36th {International}
  {Conference} on {Machine} {Learning}}}.\hskip 1em plus 0.5em minus
  0.4em\relax PMLR, May 2019, pp. 6105--6114, iSSN: 2640-3498. [Online].
  Available: \url{https://proceedings.mlr.press/v97/tan19a.html}
\BIBentrySTDinterwordspacing

\bibitem{tran_missing_2017}
L.~Tran, X.~Liu, J.~Zhou, and R.~Jin, ``Missing {Modalities} {Imputation} via
  {Cascaded} {Residual} {Autoencoder},'' 2017, pp. 1405--1414.

\bibitem{vaswani_attention_2017}
A.~Vaswani, N.~Shazeer, N.~Parmar, J.~Uszkoreit, L.~Jones, A.~N. Gomez,
  L.~Kaiser, and I.~Polosukhin, ``Attention is {All} you {Need},'' in
  \emph{Advances in {Neural} {Information} {Processing} {Systems}},
  vol.~30.\hskip 1em plus 0.5em minus 0.4em\relax Curran Associates, Inc.,
  2017.

\bibitem{wang_deep_2020}
Y.~Wang, W.~Huang, F.~Sun, T.~Xu, Y.~Rong, and J.~Huang, ``Deep multimodal
  fusion by channel exchanging,'' \emph{Advances in Neural Information
  Processing Systems}, vol.~33, 2020.

\bibitem{yeh_building_2018}
\BIBentryALTinterwordspacing
I.-C. Yeh and T.-K. Hsu, ``\BIBforeignlanguage{en}{Building real estate
  valuation models with comparative approach through case-based reasoning},''
  \emph{\BIBforeignlanguage{en}{Applied Soft Computing}}, vol.~65, pp.
  260--271, Apr. 2018. [Online]. Available:
  \url{https://www.sciencedirect.com/science/article/pii/S1568494618300358}
\BIBentrySTDinterwordspacing

\bibitem{you_image-based_2017}
Q.~You, R.~Pang, L.~Cao, and J.~Luo, ``Image-{Based} {Appraisal} of {Real}
  {Estate} {Properties},'' \emph{IEEE Transactions on Multimedia}, vol.~19,
  no.~12, pp. 2751--2759, Dec. 2017, conference Name: IEEE Transactions on
  Multimedia.

\bibitem{zeppelzauer_automatic_2018}
\BIBentryALTinterwordspacing
M.~Zeppelzauer, M.~Despotovic, M.~Sakeena, D.~Koch, and M.~Döller, ``Automatic
  {Prediction} of {Building} {Age} from {Photographs},'' in \emph{Proceedings
  of the 2018 {ACM} on {International} {Conference} on {Multimedia}
  {Retrieval}}, ser. {ICMR} '18.\hskip 1em plus 0.5em minus 0.4em\relax New
  York, NY, USA: Association for Computing Machinery, June 2018, pp. 126--134.
  [Online]. Available: \url{https://doi.org/10.1145/3206025.3206060}
\BIBentrySTDinterwordspacing

\end{thebibliography}
}

\end{document}